\pdfoutput=1

\documentclass[11pt]{article}

\usepackage[preprint]{acl}
\input{00_header.tex}

\usepackage{times}
\usepackage{latexsym}

\usepackage[T1]{fontenc}

\usepackage[utf8]{inputenc}

\usepackage{microtype}

\usepackage{inconsolata}

\usepackage{graphicx}

\title{Interactive and Expressive Code-Augmented Planning \\ with Large Language Models}

\author{
  \textbf{Anthony Z. Liu\textsuperscript{1}}\thanks{Correspondence: \href{mailto:anthliu@umich.edu}{anthliu@umich.edu}},
  \textbf{Xinhe Wang\textsuperscript{1}},
  \textbf{Jacob Sansom\textsuperscript{1}},
  \textbf{Yao Fu\textsuperscript{1}},
\\
  \textbf{Jongwook Choi\textsuperscript{1}},
  \textbf{Sungryull Sohn\textsuperscript{2}},
  \textbf{Jaekyeom Kim\textsuperscript{2}},
  \textbf{Honglak Lee\textsuperscript{1,2}}
\\
\\
  \textsuperscript{1}University of Michigan,
  \textsuperscript{2}LG AI Research 
}

\begin{document}

\maketitle

\listoffixmes

\begin{abstract}
Large Language Models (LLMs) demonstrate strong abilities in common-sense reasoning and interactive decision-making, but often struggle with complex, long-horizon planning tasks. Recent techniques have sought to structure LLM outputs using control flow and other code-adjacent techniques to improve planning performance.
These techniques include using variables (to track important information) and functions (to divide complex tasks into smaller re-usable sub-tasks).
However, purely code-based approaches can be error-prone and insufficient for handling ambiguous or unstructured data. To address these challenges, we propose \ours, an LLM planning approach that is \emph{fully code-expressive}
(it can utilize all the benefits of code) while also being \emph{dynamic} (it can flexibly adapt from errors and use the LLM for fuzzy situations).
In \ours, an LLM solves tasks by interacting with a Read-Eval-Print Loop (REPL), which iteratively executes and evaluates code, similar to language shells or interactive code notebooks, allowing the model to flexibly correct errors and handle tasks dynamically.
We demonstrate that \ours achieves strong results across various planning domains compared to previous methods.

\end{abstract}

\section{Introduction}
\label{sec:intro}

\newcommand{\fpart}[1]{\textbf{[#1]}}
\begin{figure*}[ht]
\begin{center}
\includegraphics[trim={20 55 20 60}, width=\textwidth]{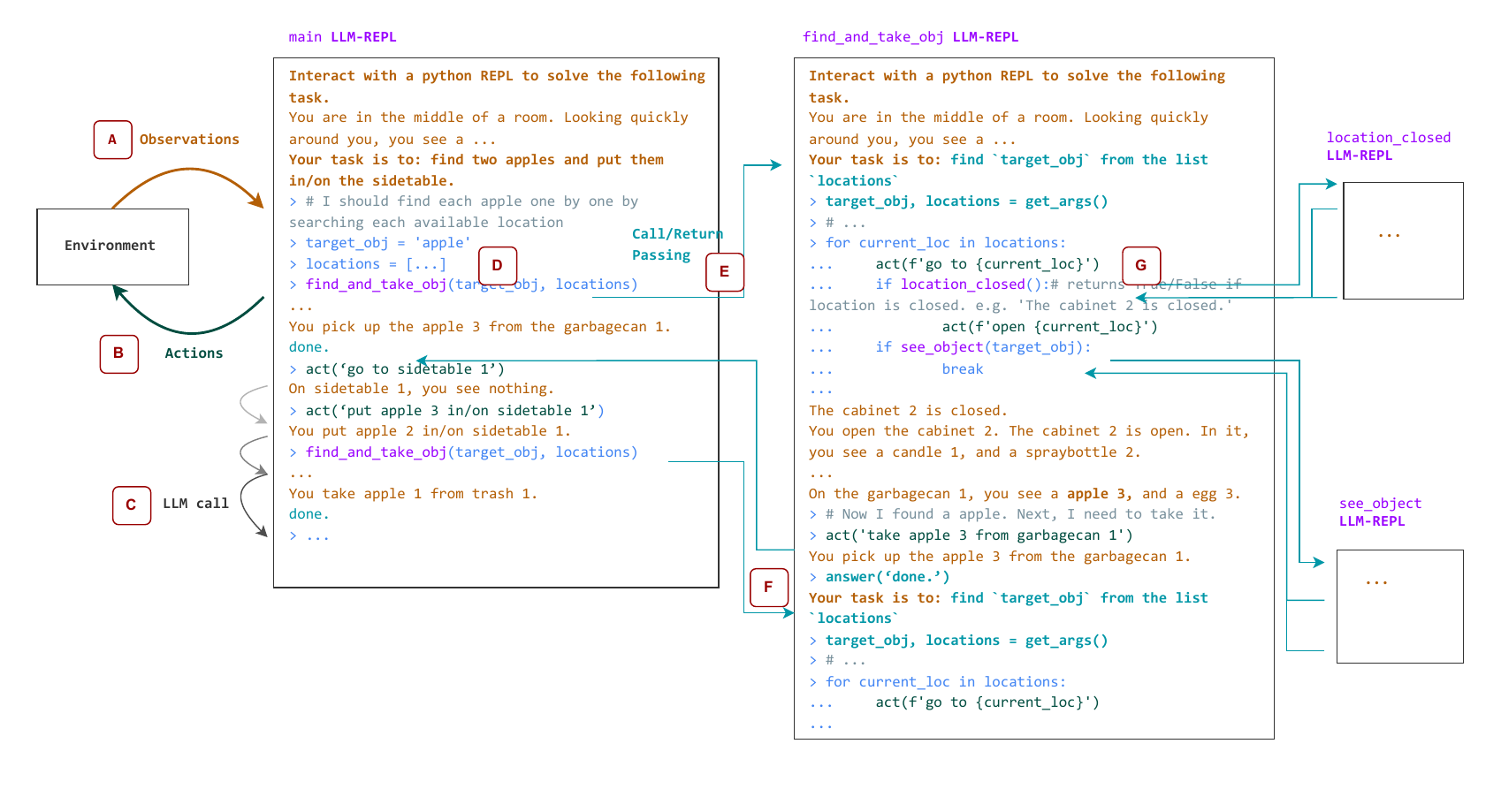}
\end{center}
\caption{\textbf{\ours} is an approach for augmenting LLM-planning by using LLMs to interact with \textbf{\oursrepls}, which are an extension of REPLs (Read-Eval-Print-Loops, e.g. language shells, code notebooks).
\fpart{A} Observations are printed out in the main \oursrepl and all child \oursrepls.
\fpart{B} \fpart{C} The LLM, which outputs code in the \oursrepl line by line, interacts with the environment by calling the \texttt{act} function. Each \oursrepl (including child) can directly act on the environment.
\fpart{D} The LLM can ``spawn'' child \oursrepls by calling undefined functions.
These functions can be used to \emph{abstract} various parts of the task (e.g.\ finding one of the necessary objects, interpreting if the current location is closed, etc.).
\fpart{E} Arguments and return values are passed between parent/child \oursrepls using the \texttt{get\_args} and \texttt{answer} functions.
\fpart{F} \oursrepls can be called multiple times, and continue output from the last \texttt{answer} statement. This allows an LLM to output \emph{consistent} and correct outputs.
\fpart{G} An LLM can combine these tools (child \oursrepls, code control flow) to express complex workflows that can solve tasks in a compact way.
}
\label{fig:overview}
\vafterfig
\end{figure*}

Large Language Models (LLMs) have shown strong capabilities in planning and decision making tasks~\citep{wang2023survey, xi2023rise}.
LLMs have strong domain-knowledge and common-sense priors~\citep{kojima2023largelanguagemodelszeroshot,brown2020languagemodelsfewshotlearners}, which are required in many of these planning tasks.
For instance, answering \emph{``What steps should I take to prepare dinner?''} requires knowledge about food preparation and the kitchen environment.
However, LLMs are imperfect and can struggle to make accurate decisions for complex and long-horizon tasks --- often making ``hallucinations'' and/or incorrect short-term decisions~\citep{kambhampati2024llms, valmeekam2023planningabilitieslargelanguage, liu2023llmpempoweringlargelanguage}.

Prior work has improved planning performance by guiding an LLM by enforcing levels of \textbf{structure} to the LLM's outputs and decisions.
A prominent method of guiding LLMs is to explicitly prompt the LLM to ``think step by step''~\citep{yao2022react, kojima2022large}, which causes the LLM to output statements to break down the given planning task into smaller tasks, or sub-tasks, and ``think'' through its decisions while planning.
Prior work has also guided LLM-planning by enforcing explicit data structures, such as dividing the task into \emph{lists} of sub-tasks and \emph{trees} of sub-tasks~\citep{yao2023tree, schroeder2024threadthinkingdeeperrecursive}, then recursively planning on these sub-tasks.

Prior work has also structured LLM decisions by using \textbf{code}, which we call here code-augmented LLM-planning.
As recent LLMs have been trained on large code bases~\citep{chen2021evaluatinglargelanguagemodels,yang2024llmwizardcodewand}, LLMs have the ability to utilize the \emph{control flow} structures of code to plan accurately.
Some works prompt an LLM to write code or pseudocode solely to help the LLM self-structure its decisions, without explictly executing code~\citep{chae2024language}.
Other works prompt an LLM to write code that is directly executed to solve the given task~\citep{gao2023palprogramaidedlanguagemodels, wang2023voyageropenendedembodiedagent}.

However, using LLM-written code also introduces the inherent limitations of code to planning.
In this work, we examine the following limitations:
(1) \emph{``Fuzzy''} sub-problems: many tasks require a planner to answer or solve some ``fuzzy'' task that is not easily solved using code. A planner may need to interpret unstructured observation data, or make decisions on possibly subjective choices (e.g.\ \emph{``purchase the item that best matches the user's request''}).
(2) \emph{Bottom-up nature} of writing code: solving tasks with code requires a planner to often solve tasks from a bottom-up manner.
This includes writing functions for sub-processes before writing code to solve the main tasks and writing case statements that must consider every possible output of a function or result of a sub-task. Generating accurate code in this manner requires precise and accurate forethought.
(3) \emph{Coding bugs}: writing accurate code in one pass is difficult, even for skilled human coders.

In this work, we seek to solve these issues by creating a \textbf{top-down} framework for code-augmented LLM-planning that is both \textbf{dynamic} and code-\textbf{expressive}.
In particular, we take inspiration from how real human developers write and prototype code, with \emph{REPL}s (Read-Eval-Print-Loops), and other interactive computing paradigms~\citep{steele1996evolution}.
A REPL, also known as a \emph{language shell}, or an interactive code notebook (e.g. IPython, Jupyter), is a coding environment wherein a programmer inputs a single statement of code, the environment executes the code and displays the output to the programmer, and then the programmer repeats the process with a new statement of code.
Using a REPL allows a programmer to write code step-by-step, see the results at each step, and correct their mistakes while interacting with the REPL.

We introduce \textbf{\ours}: an approach for solving complex planning problems (shown in \Cref{fig:overview}) by using LLMs to interact with \textbf{\oursrepls}, where the LLM writes code line-by-line, writing and calling functions to interact with the planning environment.
Specifically, \oursrepls are an extension of REPLs --- in addition to REPL functionality, code in \oursrepls can \emph{recursively ``spawn'' child \oursrepls}.
This recursive spawning allows the planner to solve tasks top-down, and can be used to abstract various subtasks and ``fuzzy'' sub-problems.

Our work makes the following contributions:
\begin{itemize}
    \item \textbf{\ours}: a novel approach for planning using \textbf{\oursrepl}s, which are an extension of REPLs (e.g.\ language shells, code notebooks).
    \oursrepls enable a LLM to make decisions in a top-down way that is both dynamic and expressive.
    \item We show that \ours achieves strong performance in challenging text-based sequential decision making environments ALFWorld~\citep{shridhar20alfworld} and WebShop~\citep{yao2022webshop}.
    We also show \ours achieves strong performance on a novel real-world web navigation task that requires handling complex web observations.
    \item We test the robustness of \ours by conducting ablations --- testing the code correction/prediction ability by giving the LLM faulty demonstrations.
\end{itemize}

\section{Approach}

\subsection{Problem Statement}

In a typical interactive planning task, a large language model agent (LLM-agent) is given a task description $t$ and must interact with an environment for a number of timesteps to accomplish the task $t$. At each timestep $i$, the agent is given a language observation $o_i$ of the current state of the environment and can predict a language action $a_i$ which can affect the environment (for the agent to accomplish $t$).
To help the LLM-agent, it is given $k$-trajectories (or demonstrations) $\{\tau_1, \ldots, \tau_k\}$ of solving some prior tasks that are created by an expert. The LLM-agent must follow the demonstrations to solve the current task $t$.

A simple way to apply the LLM is to simply model the next action using the language-modeling probabilities of the LLM.
However, many real-world tasks are complex --- they may require solving multiple smaller sub-tasks, and long-horizon --- they require many action predictions to solve the task.
Enforcing structure, such as writing \emph{code}, to the LLM-agent's decisions, can be used to address these issues.

In this work, we propose a novel method for using code to augment LLM-agents, which addresses the limitations of using code with planning, described in~\Cref{sec:intro}: (1) ``Fuzzy'' sub-problems, (2) planning bottom-up, and finally (3) coding errors.

\begin{figure*}[ht]
\vspace{-5pt}
\centering
\small

\newcommand{\lcodewidth}{0.44\textwidth}
\newcommand{\rcodewidth}{0.55\textwidth}
\begin{subtable}[t]{\textwidth}
\begin{tabular}{p{\lcodewidth}p{\rcodewidth}}
\texttt{filter\_search()} REPL & \texttt{filter\_page()} REPL \\
\hline
\vspace{-10pt}
\begin{lstlisting}[language=python, linewidth=\lcodewidth]
Task: Find any matching items in the search results.
>>> description = get_args()
>>> matching = []
>>> for i in range(5):
...     matching.extend(filter_page(description))
...     act('click [Next >]')
...
>>> answer(matching)
\end{lstlisting}
&
\vspace{-10pt}
\begin{lstlisting}[language=python, linewidth=\rcodewidth]
Task: Find any matching items on the current page
>>> description = get_args()
>>> item_links = parse_items()
>>> matching = []
>>> for item in range(item_links):
...     act(f'click [{item}]')
...     if item_matches(item, description):
...         matching.append(item)
...     act(f'click [< Back]')
...
>>> answer(matching)
\end{lstlisting}
\end{tabular}
\end{subtable}

\vspace{-12pt}

\begin{subtable}[t]{\textwidth}
\centering
  \rowcolors{2}{gray!10}{white}
\begin{tabular}{lll}
Actions         & \texttt{filter\_search()} Trace                              & \texttt{filter\_page()} Trace \\
\hline
                & $\ell_2 \gets \texttt{LLMQuery}()$ &       \\
                & \texttt{description = get\_args()} & \\
                & $\ell_3 \gets \texttt{LLMQuery}()$ &       \\
                & \texttt{matching = []} & \\
                & $\ell_7, \ldots, \ell_7 \gets \texttt{LLMQuery}()$ &       \\
                & \texttt{for i in range(5):} & \\
                & \textcolor{red}{\texttt{NameError}}\texttt{(filter\_page)} & \\
                & create REPL: subtask $\gets \texttt{LLMQuery}()$ & \\
                & \quad\texttt{matching.extend(} & \\
                & \qquad\texttt{filter\_page(description))} & \\
                &               & $\ell_2 \gets \texttt{LLMQuery}()$ \\
                &               & \texttt{description = get\_args()} \\
                &               & $\ell_3 \gets \texttt{LLMQuery}()$ \\
                &               & \textcolor{red}{\texttt{NameError}}\texttt{(parse\_items)} \\
                &               & create REPL: subtask $\gets \texttt{LLMQuery}()$ \\
                &               & \texttt{item\_links = parse\_items()} \\
                &               & $\ell_4 \gets \texttt{LLMQuery}()$ \\
                &               & \texttt{matching = []} \\
                &               & $\ell_5, \ldots, \ell_9 \gets \texttt{LLMQuery}()$ \\
                &               & \texttt{for item in range(item\_links):} \\
\texttt{act('click [\emph{X}]')} &               & \quad\texttt{act(f'click [\{item\}]')}  \\
                &               & \quad\textcolor{red}{\texttt{NameError}}\texttt{(item\_matches)} \\
$\ldots$ &               & \quad$\ldots$  \\
\texttt{act('click [< Back]')} &               & \quad\texttt{act(f'click [< Back]')}  \\
$\ldots$ &               & \quad$\ldots$  \\
                &               & \texttt{answer(matching)} \\
\texttt{act('click [Next >]')} & \quad\texttt{act('click [Next >])'} & \\
$\ldots$ & $\ldots$      & $\ldots$  \\
& \texttt{answer(matching)} & \\
\end{tabular}
\vspace{-8pt}
\end{subtable}

\caption{A toy example of context passing that is possible in \ours.
In the toy example, the task is parse all items on a search page that match a given description.
We show a sample generated code in the two code snippets above, where the agent splits the task into 3 different sub-tasks:
(1) \texttt{filter\_page}, parsing any matching items on the current page,
(2) \texttt{parse\_items}, parse any item links on the current page,
and
(3) \texttt{item\_matches}, determine if the current item page matches the description.
}
\label{tab:cxt-pass}
\vafterfig
\end{figure*}

\subsection{\oursrepls}

We base our approach on the concept of a REPL (Read-Eval-Print-Loop), also known as a language shell or code notebook, is a programming environment that tracks program state (global/local variables) at each timestep
where (1) a user inputs a statement of code (variable assignment, loop, function definition),
(2) this line of code is \emph{evaluated}, to obtain the next program state,
(3) the output of the previous line of code is shown to the user,
and finally (4) repeat step (1).

A REPL is useful for task solving in a \emph{dynamic} manner --- results (and possible errors) are immediately reflected in the output of the program state. Then, a user (or LLM), can simply write another statement of code that can correct their previous mistake, without re-writing the entire program.

We extend the idea of using a REPL \emph{recursively} --- where any REPL can spawn \emph{another} REPL in a top-down manner, in order to solve any (possibly fuzzy) sub-task --- a ``subtask'' REPL.
We call this idea an \textbf{\oursrepl}. 
In \Cref{sec:ablations}, we show experimentally adding recursive spawning is crucial for solving tasks.

To create a \textbf{\oursrepl}, we simply add pre-defined functions, or \emph{primitives}, to the program state that can \emph{recursively spawn} child \oursrepls, and manage context-passing between parent-child \oursrepls.
\begin{itemize}
    \item \texttt{[subtask](args)}: spawns an \oursrepl with the name \texttt{[subtask]}.
    The LLM is then queried using the \texttt{[subtask]} REPL until context is passed back to this REPL or another child REPL is spawned.
    If the \texttt{[subtask]} \oursrepl has previously been spawned, then execution history will \emph{continue} from previous. Otherwise, a task description for the \oursrepl will be queried, and a new \oursrepl will be created. Note the new child \oursrepl does not share any variable state with the parent, so the following primitives can be used to pass context.
    \item \texttt{get\_args()}: returns the arguments passed from the parent of the current \oursrepl.
    \item \texttt{answer(a)}: passes and returns the value \texttt{a} back to the parent \oursrepl. Execution is passed back to the parent.
\end{itemize}
Pseudo-code for how a practical implementation of \oursrepls is given in \Cref{sec:impl-details}.

\subsection{\ours}

\textbf{\ours} is simply planning and decision-making through collection of \oursrepls.
As all the planning and actions can be done in a \oursrepl, \ours is code-\emph{expressive}.

To execute \textbf{\ours}, we can adapt an \oursrepl by adding more primitive functions for interacting with the environment. 
\begin{itemize}
    \item \texttt{act(a)}: passes and executes the action \texttt{a} in the environment.
    \item \texttt{get\_obs()}: passes the recent observation as a string. This may be used by an planner to execute some string processing code when an observation is too large (e.g.\ a whole web page).
\end{itemize}
 For each timestep $i$, we run the \oursrepl until the \texttt{act(a)} function is called, then pass in any observations as needed to the in-code-environment-outputs of the \oursrepls.
 We demonstrate how LLM calls, code, and actions from \texttt{act} are executed across different parent-child \oursrepls in \Cref{tab:cxt-pass}.

\paragraph{$k$-shot \ours.}
To use \ours in multiple planning tasks, we keep a \emph{global REPL pool}, a set of spawned REPLs from previous task executions and demonstrations.
Note that an \oursrepl, when called or spawned, \emph{re-uses} the code/output history.
In this way, an LLM can perform in-context-learning from previous tasks and demonstrations to output correct and consistent code.

\section{Experiments}

We evaluated \ours against baselines on various language-based environments.
Each of these environments is challenging --- to be successful, an agent must be able to understand and predict long-horizon complex action sequences,
and have enough language-understanding to interpret the language observations and actions.
We show that \ours is an effective balance in using code for planning --- using code to structure complex, repeating, and language-understanding sub-tasks.

\subsection{Environments}

We tested on the following environments:
\paragraph{ALFWorld.}
ALFWorld~\citep{shridhar20alfworld} is a text-based simulated-household embodied agent environment, where an agent must accomplish 6 different types of tasks related to navigating and interacting with a household, e.g. searching the kitchen for an apple, heating it, then placing it on a table.
\paragraph{WebShop.}
WebShop~\citep{yao2022webshop} is an interactable e-commerce environment, where an agent is given a target product description---in which a human annotator described a category, key attributes, and price limit---and the agent must navigate the e-commerce site by searching for and buying a product that matches the target description. An agent's success is measured based on how many of the purchased item's attributes match with the ground truth attributes.

In WebShop, the search page is configured to return the top $k$ results from the search query.
Prior approaches have set $k=3$,
and tested strategies where the best item is selected only from the first page (Top-3).
We additionally test a setting where $k=10$.
In this setting, we tested the strongest baseline, THREAD~\citep{schroeder2024threadthinkingdeeperrecursive} ---an LLM approach that uses a text-based divide-and-conquer strategy--- by (1) using the same Top-3 strategy, and (2) using a Top-20 strategy --- searching twice and selecting the best among 20. Details included in~\Cref{sec:thread-mods,sec:webshop-mini}.

\paragraph{Real-World Web Tasks.}
To test the scalability of \ours, we further designed a set of real-world web navigation tasks on actual websites. These tasks are difficult as actual websites can be extremely large (4k-20k tokens, compared to $\leq 500$ tokens per web-page in WebShop).
We tested the agents on distributions of tasks where the goal is to navigate an E-commerce website and add item(s) that match a given description to the cart.
We divided the tasks into a ``simple'' category, where the agent only needs to add an item of any description (e.g.\ add any laptop to the cart from \texttt{bestbuy.com}) and a ``complex'' category, where the agent must add as many items as it can find that match a specific description (e.g.\ loop through each search result and add every printer that is capable of printing at least 12 pages per minute to the cart).
The full task descriptions and examples of observations and actions can be found in \Cref{sec:rww-details}.

\subsection{Results}

We find that elements of \emph{top-down recursive, dynamic}, and \emph{code-expressivity} are important for approaches to score a high success rate on all environments (top-down recursive and dynamic for ALFWorld, dynamic and code-expressivity for WebShop and Real-World Web).
For example, the baseline THREAD~\citep{schroeder2024threadthinkingdeeperrecursive} has top-down recursive and dynamic elements, but not code-expressivity, has high success rate on ALFWorld and $k = 3$ WebShop, but fails to scale in $k = 10$ WebShop and Real-World Web.

\begin{table}[t]
\centering
\tablesize
\begin{tabular}{llc}\toprule
\multirow{2}{*}{\makecell{External \\ Memory}}
 & & \multirow{2}{*}{\makecell{ALFWorld \\ SR (\%)}} \\
 & Method & \\
\midrule
\multirow{4}{*}{Yes} & Reflexion$^1$  & 76.1 \\
& AdaPlanner$^2$ & 82.8 \\
& RAP$^3$  & 85.8 \\
& AutoGuide$^4$ & 79.1 \\
\cmidrule(lr){1-3}
\multirow{4}{*}{No} & ReAct$^5$   & 53.7 \\
& ADaPT$^6$ & 82.1 \\
& THREAD$^7$  & 95.5 \\
& \ours & \textbf{97.0} \\
\bottomrule
\end{tabular}
\caption{The average success rate (SR (\%)) of running each method on ALFWorld with GPT-3.5-instruct~\cite{openai2023gpt35}.
We note that methods 1-4 use external memory --- they save information from previous test tasks to new test tasks.
$^1$~\cite{shinn2023reflexion}
$^2$~\cite{sun2023adaplanneradaptiveplanningfeedback}
$^3$~\cite{kagaya2024rapretrievalaugmentedplanningcontextual} 
$^4$~\cite{fu2024autoguideautomatedgenerationselection}
$^5$~\cite{yao2022react}
$^6$~\cite{prasad2024adaptasneededdecompositionplanning}
$^7$~\cite{schroeder2024threadthinkingdeeperrecursive}
}
\label{tab:alfworld}
\vafterfig
\end{table}

\begin{table*}[t!]
\centering
\tablesize
\begin{tabular}{lllllcc}\toprule
& & & & & \multicolumn{2}{c}{WebShop} \\
\cmidrule(lr){6-7}
Model & Setting & Strategy & External\ mem.\ & Method & SR(\%) & Score(\%) \\
\midrule
\multirow{9}{*}{GPT-3.5-instruct} & \multirow{9}{*}{Page $k=3$} & \multirow{9}{*}{Top-3} & \multirow{4}{*}{Yes}
& Reflexion & 38 & 64.4 \\
& & & & LATS$^1$   & 40 & 76.0 \\
& & & & RAP   & 48 & 76.1 \\
& & & & AutoGuide   & 46 & 73.4\\
\cmidrule(lr){4-7}
& & & \multirow{5}{*}{No} & ReAct   & 37 & 59.5 \\
& & & & ADaPT   & 44 & 60.0 \\
& & & & TDAG$^2$   & 45 & 64.5 \\
& & & & THREAD & \textbf{49} & \textbf{76.3} \\
& & & & \ours & 47 & 74.2\\
\midrule
\multirow{3}{*}{GPT4o-mini} & \multirow{3}{*}{Page $k=10$} & Top-3 & \multirow{3}{*}{No}
& THREAD$^*$ & 21 & 42.1 \\
& & Top-3 & & \ours & 37 & 69.9 \\
& & Top-20 & & \ours & \textbf{52} & \textbf{77.1} \\
\bottomrule
\end{tabular}
\caption{The success rate (SR) and average score of each model-method-setting on Webshop.
For the $k=3$ setting, models were run using GPT-3.5-instruct and with a top-3 strategy (examine only the top-3 items).
For the $k=10$ setting, we ran THREAD and \ours using GPT4o-mini~\cite{openai2023gpt4}, and showed the performance of each the viable strategy. $^*$ note that we made modifications to run THREAD as intended, described in~\Cref{sec:thread-mods}.
$^1$\cite{zhou2024languageagenttreesearch}
$^2$\cite{wang2024tdagmultiagentframeworkbased}
}

\label{tab:webshop}
\vafterfig
\end{table*}

\begin{table}[t]
\centering
\tablesize
\begin{tabular}{lS[table-format=2.1]S[table-format=2.1]}
\toprule
& \multicolumn{2}{c}{Real-World Web} \\
& \multicolumn{2}{c}{(\% of Expert Score)} \\
\cmidrule(lr){2-3}
Method & {Simple} & {Complex} \\
\midrule
ReACT   & \textbf{86.7} & 17.6 \\ %
THREAD   & 13.3 & 0.0 \\ %
\ours & \textbf{86.7} & \textbf{39.6} \\ %
\bottomrule
\end{tabular}
\caption{We tested the baselines ReACT and THREAD against \ours in the Real-World Web Tasks with GPT4o-mini. For each method, we present the percentage of the expert score reached.}
\label{tab:rwweb}
\vafterfig
\end{table}

\begin{figure*}[t]
\begin{center}
\includegraphics[trim={20 65 20 60}, width=\textwidth]{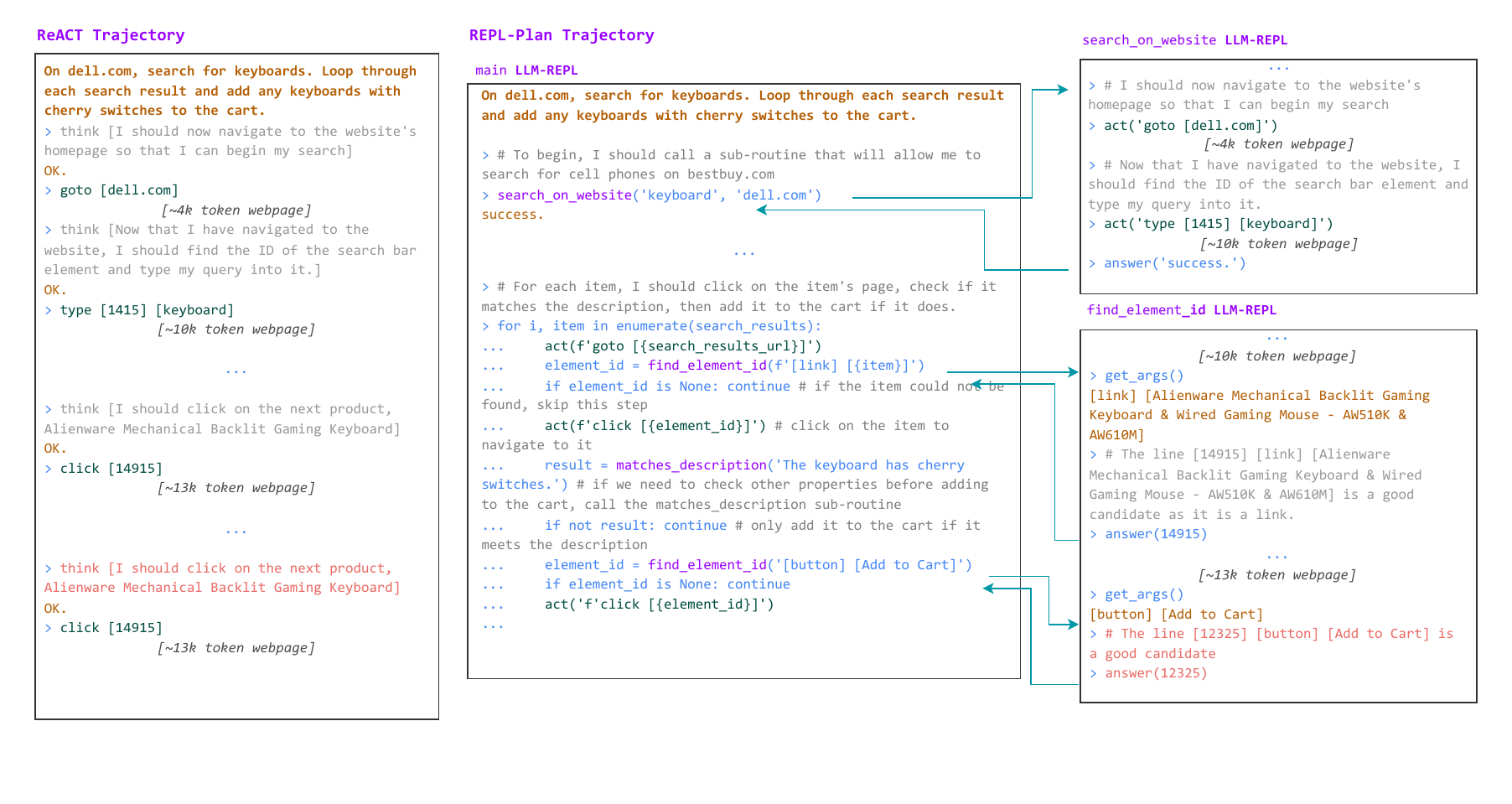}
\end{center}
\caption{For a qualitative analysis, we include truncated versions of trajectories from \ours and the baseline ReACT on a real-world web loop-like task. In this task (\Cref{sec:rww-details}), the agent is shown long web pages (4k-15k tokens long), and must interact with the page using element IDs (labeled with integer IDs). 
In the trajectories, (1) \ours is able manage large observations and long prompt contexts by sub-dividing the tasks into different \oursrepls.
And (2), we find that in both trajectories, ReACT and \ours both run into hallucination errors from GPT4o-mini (highlighted in red). On the left, in ReACT, the LLM gets ``lost'', and re-checks a product it already checked before. This causes the ReACT agent to loop infinitely. On the right, in \ours, agent hallucinates a link element ID. However, due to code in the main loop \ours mitigates the effect of the hallucination --- the agent clicks the wrong element ID, but still continues to search for candidate products.
}
\label{fig:rww-qualitative}
\vafterfig
\end{figure*}

\paragraph{ALFWorld.} We show the results of testing approaches on ALFWorld in \Cref{tab:alfworld}.
Our approach, \ours, achieves the highest success rate (SR) of 97.0\% compared to the best baseline, THREAD~\citep{schroeder2024threadthinkingdeeperrecursive}, with 95.5\% SR.

We find that \emph{top-down recursive} and \emph{dynamic} approaches are crucial to solving this task, which ours and THREAD both have. Both \ours and THREAD effectively divide-and-conquer various sub-tasks of ALFWorld (e.g.\ dividing a task into (1) finding an apple, (2) heating it, then (3) placing it on the table),
and are able to effectively change plans if there are environment issues (e.g. object is misplaced or mis-referenced).

\paragraph{WebShop.} We show the results of various approaches on different LLMs, settings, and strategies on WebShop in \Cref{tab:webshop}.

For simple strategies such as Top-3, i.e. simply searching and then comparing the Top-3 items, many baselines score a high ($\geq 46$\%) success rate: RAP, AutoGuide, THREAD, and \ours, with THREAD reaching the the highest SR of 49\%.
However, this strategy is simplistic and limited --- trajectories can be completed within 10 actions, and success primarily relies on the LLM's ability to understand web-page observations.

By testing on the Page $k=10$ setting with GPT4o-mini on \ours and the strongest baseline THREAD, we find that a combination of \emph{top-down recursive} and code-\emph{expressive} approaches are valuable in attaining a high SR in this setting and being able to successfully execute a Top-20 strategy.

\ours, using loops and variables (that can be passed across \oursrepls), is able to reach a 52\% SR with GPT4o-mini.

\paragraph{Real-World Web Environment.}

We show the results of running ReACT, THREAD, and \ours in the Real-World Web Environment in \Cref{tab:rwweb}.
We find that each agent is able to solve tasks from the Simple distribution (tasks outlined in ~\Cref{sec:rww-details}).
These tasks are short horizon and can be completed in 5 actions. However, they still require the agent to be able to manage and interpret the long web page observations, which can be 4k-20k tokens long.
Despite multiple attempts to improve THREAD described in ~\Cref{sec:thread-mods}, we found THREAD could not effectively plan when there are long observation contexts.
\ours is able to more effectively break down the observations using full code expressitivity --- scoring 39.6\% of the expert score on the complex tasks, where the baselines ReACT and THREAD score 17.6\% and 0.0\% respectively.
We also give a qualitative analysis of this result in \Cref{fig:rww-qualitative}.
In this analysis, we show why \ours is able to (1) effectively manage large observations through code and \oursrepl spawning, and (2) mitigate the effects of LLMs hallucinating using code.

\subsection{Ablations}
\label{sec:ablations}

\begin{table*}[ht]
\centering
\tablesize
\begin{tabular}{llcc}
\toprule
& & \multicolumn{2}{c}{WebShop sample SR (\%, $n = 25$)} \\
\cmidrule(lr){3-4}
Method & Ablation & {GPT-3.5-instr.} & {GPT-4o-mini} \\
\midrule
\multirow{4}{*}{\ours}
& Full Model & 52 & 44 \\ %
& (1) Buggy Demo. & 52 & 40 \\ %
& (2) No-Subtask-REPLs & 24 & 20 \\ %
& (3) Zero-shot Subtask-REPL & 28 & 16 \\ %
\bottomrule
\end{tabular}
\caption{We tested \ours on the WebShop Page $k=3$ setting with various corruptions to the demonstrations.
(1) Buggy Demo.: We added minor bugs (non-escaped strings, missing variables) to the demonstration.
(2) No-Subtask-REPLs: We tested a version of \ours (and demonstrations) without recursive spawning.
(3) Zero-shot Subtask-REPL: We removed one of the Subtask-REPLs from the demonstration, such that the agent needed to zero-shot infer code.
}
\label{tab:ablations}
\vafterfig
\end{table*}

We tested \ours's ability to run under imperfect conditions in the page $k=3$ WebShop environment, using the Top-3 strategy.

\paragraph{Buggy Demonstrations.}
In order to test the error-correcting abilities of \ours, we injected minor coding mistakes
into the demonstrations and tested the performance.

Specifically, we injected two errors: (1) we saved variable descriptions using quotes: 
\texttt{description = "..."}. 
This would cause errors when there are quotes in the desired description contained quotes: e.g.\ \texttt{" 12" ..."},
and (2) we removed key variable definitions that are used later in the demonstration.
We found that the LLM is relatively robust to these errors, only failing one more task on GPT-4o-mini than without any errors. We show a small sample of these corrections in \Cref{sec:bug_correct}.

\paragraph{No-Subtask-REPLs.}
Next, we tested how the top-down decomposition of tasks affects the performance of \oursrepl.
We created a ``No-Subtask-REPL'' version of REPL-Plan, the agent aims to solve the task without spawning any child \oursrepls.%

We find this strategy is conceptually similar to a code-augmented version of prior approaches, such as ReACT~\citep{yao2022react}.
However, given that this approach cannot de-compose tasks, the REPL code/observation history becomes large and untenable for the agent to solve WebShop tasks.
The performance decreases by around half: $52\% \to 24\%$ and $44\% \to 20\%$ respectively for GPT-3.5-instr.\ and GPT-4o-mini.

\paragraph{Zero-shot Subtask-REPL.}
As $k$-shot demonstrations are given as a pool of \oursrepls which are used in \ours, we tested the performance under the condition that one of those \oursrepls was removed from this pool, and the LLM must zero-shot infer both the task description and code writing for this \oursrepl.

To do this, we removed the \texttt{get\_requirements} \oursrepl, which is designed to examine the item on the current page, and return an integer count of the number of matching attributes.
This is a crucial function that is necessary for completing the task and called multiple times.

We found that in half of the trials the LLM is able to correctly zero-shot infer a subtask description and code for the \oursrepl, with a final performance of $28\%$ (from $52\%$) and $16\%$ (from $44\%$) respectively for GPT-3.5-instr.\ and GPT-4o-mini.
When the LLM incorrectly infers this subtask, it often causes the agent to enter a failure loop, causing the agent to fail the task overall.
We analyze success and failure cases more closely and give qualitative analysis in \Cref{sec:zs-repl-details}.
Although zero-shot task/code inference is not the focus of this work, this indicates an extension work that could be achieved by fine-tuning LLMs to understand REPLs and \oursrepls beyond in-context learning.

\section{Related Work}
\paragraph{LLM-Agents.}
With the increasing capabilities of LLMs, several works have proposed robust agents that interact with text-based environments (LLM-agents). These works show promising results in challenging benchmarks such as embodied agent tasks~\citep{shridhar20alfworld} and web navigation~\citep{yao2022webshop, zhou2023webarena}.
A number of these LLM-agents, based on works by ~\citet{yao2022react} and~\citet{shinn2023reflexion},
take in a sequence of text observations-action pairs as history and generate single actions at each time-step.
However, as the LLMs tend to hallucinate incorrect actions, and incorrectly plan ahead, several works, including \ours, have been proposed that incorporate \emph{code}-augmentation to LLM-agents, and recursively dividing tasks for an LLM-agent.

\paragraph{Code-Augmented LLMs.}
Current LLMs (without guidance) suffer from long context understanding~\citep{li2024longcontextllmsstrugglelong} and logical and arithmetic mistakes~\citep{wei2023chainofthoughtpromptingelicitsreasoning}. To alleviate these problems, recent works have used external tooling such as code.
For example, Program-Aided Language Models (PAL)~\citep{gao2023palprogramaidedlanguagemodels}
generate programs and run a Python interpreter to get solutions for mathematical, symbolic, and algorithmic reasoning problems.
LLMs and code have also been used in robotics to generate robust policies~\citep{liang2023codepolicieslanguagemodel}.
Compared to these works, our work \ours focuses on the interactive decision-making setting, 
and focuses on how we can fix the problems of running code in these settings by creating an approach that is fully interactive (the LLM interacts with a \oursrepl to code) and top-down recursive (with spawning child \oursrepls).
CodeAct and Mint~\citep{wang2024executablecodeactionselicit,wang2023mint} are works that proposes an interactive LLM-code generating paradigm where the agent interacts with a human user to solve a task using code. However, compared to \ours, these works are not designed for the LLM-agent space, and also do not split interactive code in a recursive manner.
AdaPlanner~\citep{sun2023adaplanneradaptiveplanningfeedback} also focuses on the planning setting by writing and refining code, but is not interactive, as a code plan must be written entirely in one pass.

\paragraph{Decomposing Tasks.}
In the few-shot in-context learning setting, LLMs can struggle with increased task complexity.
To deal with this, ~\citet{khot2023decomposedpromptingmodularapproach} have proposed to decompose challenging tasks to simpler sub-tasks and solve these individually. 
ADaPT~\citep{prasad2024adaptasneededdecompositionplanning} recursively decomposes complex sub-tasks as-needed to adapt to LLMs' capability and task complexity.
Prior works ~\citep{wang2023voyageropenendedembodiedagent,liu2024skillact} have decomposed tasks into re-usable code blocks, or prompts, which can be re-used in new tasks.
THREAD~\citep{schroeder2024threadthinkingdeeperrecursive} is a work that recursively divides LLM context into \emph{thread}, which each individually solve some sub-task of the whole.
Thread also contains partial Python code executability, by saving variables and computing one-line Python statements.
\ours is different from THREAD: (1) \ours is fully code-expressive, any variables can be passed to other REPLs, loops can be run, etc. (2) \ours handles recursive sub-task spawning in a code-expressive way (using functions). In THREAD, these sub-tasks are spawned ``anonymously'', and cannot be re-used.
These two differences allow \ours to efficiently handle program contexts in many planning situations.

\section{Conclusion}
In this work, we proposed REPL-Plan, a code-augmented planning approach for Large Language Models (LLMs) that is both dynamic and code-expressive, enabling effective top-down planning. By extending the Read-Eval-Print Loop (REPL) paradigm, we create LLM-REPLs, which allow LLMs to interactively generate code, recursively spawn child LLM-REPLs, and dynamically adapt to complex tasks. Through evaluations on diverse planning benchmarks, REPL-Plan demonstrates superior performance compared to baseline methods, showcasing its ability to handle complex, long-horizon tasks with flexibility and robustness.

\section{Limitations}

One limitation of this work is the abilities of Large Language Models (LLMs) to generalize beyond $k$-shot settings --- particulary when using \ours. As we showed in the ablation experiments in \Cref{tab:ablations}, LLMs have some capabilities of generalizing beyond the $k$-demonstrations (when the $k$-demonstrations contain partially buggy code, zero-shot subtask-REPL inference). However, beyond this, such as writing zero-shot code for new \oursrepl-code for more significant subtasks, and generalizing from improperly structured code (No-Subtask-REPLs),
it can be difficult for the LLMs to successfully use \ours.
Future work could address this by examining how to demonstrate \ours under out of distribution settings, and also examining how to adapt LLMs for REPL-coding in a more manner beyond in-context learning (e.g.\ fine tuning LLMs to write code for REPLs).

Another limitation is the efficiency of some use-cases of \ours. In many scenarios, a subtask can be solved using a pure code approach --- an LLM writes code for an \oursrepl that can solve the subtask without intervention. Future work could address this by having a mechanism for determining when code written by the LLM is independent of LLM decisions, and ``automate'' subtasks if possible.

\section*{Acknowledgments}

This work was supported in part by funding from LG AI Research and NSF Graduate Research Fellowship.

\bibliography{refs}

\newpage

\onecolumn
\appendix
\section{Appendix}

\subsection{LLM-REPL Implementation}
\label{sec:impl-details}

\renewcommand{\KwSty}[1]{\textnormal{\textcolor{blue!90!black}{\ttfamily\bfseries #1}}\unskip}
\renewcommand{\ArgSty}[1]{\textnormal{\ttfamily #1}\unskip}
\SetKwComment{Comment}{\color{green!50!black}// }{}
\renewcommand{\CommentSty}[1]{\textnormal{\ttfamily\color{green!50!black}#1}\unskip}
\newcommand{\var}{\texttt}
\newcommand{\FuncCall}[2]{\texttt{{\bfseries #1}(#2)}}
\newcommand{\Pseudo}[1]{\textnormal{\ttfamily\color{purple!70!black}/* #1 */}\unskip}
\SetKwProg{Function}{def}{}{}
\renewcommand{\ProgSty}[1]{\texttt{\bfseries #1}}
\SetKwProg{Class}{class}{}{}
\SetKwProg{Try}{try}{}{}
\SetKwProg{Except}{except}{}{}
\begin{algorithm*}[ht]
    \caption{LLM-REPL Pseudocode}
    \label{alg:llmrepl}
    \Class{LLMREPLFunction}{
        \Function{\_\_init\_\_($\var{task\_description}$)}{
            $\var{task} \gets \var{task\_description}$\;
            $\var{history} \gets []$ \Comment{REPL code/output history}
            $\var{locals} \gets \{\var{get\_args, answer, act, get\_obs}\}$
                \Comment{Insert REPL primitives to local state}
            $\var{pool} \gets$ \Pseudo{global REPL pool}\;
        }
        \Function{\_\_call\_\_($\var{*args}$)}{
            \While {True} {
                \If {\Pseudo{need new REPL block}}{
                    $\var{code} \gets$ \FuncCall{QueryLLM}{$\var{task, history}$}\;
                }
                \Try{}{
                    \FuncCall{exec}{$\var{code}, \var{locals}$}
                        \Comment{Return if answer is called}
                    $\var{history.append}$(\Pseudo{format code, stdout, stderr})
                        \Comment{only reached if code finishes without interrupt}
                }
                \Except{$\var{NameError}$}{
                    \Comment{When LLM calls an unseen function}
                    $\var{fname} \gets$ \Pseudo{get undefined variable name}\;
                    \Pseudo{If REPL not in global REPL pool}\;
                    $\var{subtask} \gets$ \FuncCall{QueryLLM}{$\var{fname, history}$}\;
                    $\var{locals[fname]} \gets$ \FuncCall{LLMREPLFunction}{$\var{subtask}$}\;
                }
                \Pseudo{Return to code state after last interrupt}
            }
        }
    }
\end{algorithm*}

Implementing an LLM-REPL, as described in~\Cref{alg:llmrepl}, can be accomplished in various approaches.
The primary implementation challenge is how to handle child LLM-REPL spawning --- whenever there is an undefined function call, an LLM-REPL should be created and run for that undefined function. 
Given the undecidable nature of determining whether a function or variable is undefined~\cite{sipser1996introduction}, the most flexible approach is to handle any undefined variable/function errors, \texttt{NameError}s in Python, during runtime.

One approach is to modify a code interpreter (e.g.\ the Python interpreter) to automatically intercept these errors and continue execution.
Continuing execution between parent and child LLM-REPLs can be synced by running different threads or co-routines with various concurrency synchronization primitives.

We implemented LLM-REPL without modifying the interpreter or running any concurrency routines by saving execution state for each REPL loop iteration and re-running code when LLM-REPLs are spawned. When a \texttt{NameError} is encountered, we create an LLM-REPL and add it as a variable to the last saved execution state. Finally, we re-run the code from this last execution state.
However, an issue with this approach is multiple context sensitive functions (other LLM-REPLs, \texttt{act}) can be called again after an interrupt.
For example, consider the execution trace from \Cref{tab:cxt-pass}. After the \texttt{NameError(count\_even)}, \texttt{act(i*2+1)} will be called again after re-running the code.
Avoiding incorrect and duplicate computation can avoided by keeping a function call counter for each LLM-REPL, then when a context-sensitive function called more than once, the return value is returned using a cache.
These call counters must be updated and synced between parent-child LLM-REPLs after each context-passing function.

Another implementation detail is how to differentiate between a code statement that intends to spawn a new child LLM-REPL, and a code statement that makes a mistake by forgetting to define a variable.
For example, consider the following lines of code:
\begin{enumerate}
    \item \texttt{[id for id, price in id\_to\_price.items() for price < max\_price]}
    \item \texttt{[id for id, price in id\_to\_price.items() for price < get\_max\_price()]}
\end{enumerate}
In the case that \texttt{max\_price} and \texttt{get\_max\_price} are undefined, both lines would raise 
a \texttt{NameError}. In both cases, in our framework, an LLM-REPL will be created and named \texttt{max\_price} and \texttt{get\_max\_price} respectively.
However, in the implementation of an LLM-REPL, we override \emph{any non-function method} (such as comparators and string representations) to raise a custom error: \texttt{REPLNameError}.
Because of this, the first line of code 1.\ will result in a \texttt{REPLNameError} raised where the LLM will be given feedback that it forgot to define the \texttt{max\_price} variable. The second line of code 2.\ will jump to the \texttt{get\_max\_price()} LLM-REPL execution.

We have attached our implementation for LLM-REPLs with example test cases in the supplementary material.

\begin{figure*}[ht]
\vspace{-5pt}
\centering
\small

\begin{subtable}[t]{\textwidth}
\begin{tabular}{p{0.44\textwidth}p{0.44\textwidth}}
\texttt{main()} REPL & \texttt{count\_even()} REPL \\
\hline
\vspace{-10pt}
\begin{lstlisting}[language=python, linewidth=0.48\textwidth]
Task: Count to 4.
>>> for i in range(2):
...     act(i*2+1)
...     count_even()
...
\end{lstlisting}
&
\vspace{-10pt}
\begin{lstlisting}[language=python, linewidth=0.48\textwidth]
Task: Count only evens to 4.
>>> for i in range(2):
...     act((i+1)*2)
...     answer(f'Counted {i*2}.')
...
\end{lstlisting}
\end{tabular}
\end{subtable}

\vspace{-12pt}

\begin{subtable}[t]{\textwidth}
\centering
  \rowcolors{2}{gray!10}{white}
\begin{tabular}{lll}
Actions         & \texttt{main()} Trace                              & \texttt{count\_even()} Trace \\
\hline
                & $\ell_2, \ldots, \ell_5 \gets \texttt{LLMQuery}()$ &       \\
                & \texttt{for i in range(2):} & \\
\texttt{act(1)} & \quad\texttt{act(i*2+1)} & \\
                & \textcolor{red}{\texttt{NameError}}\texttt{(count\_even)} & \\
                & create REPL: subtask $\gets \texttt{LLMQuery}()$ & \\
                & add \texttt{count\_even} REPL to variables & \\
                & \emph{``rewind''} state back to $\ell_4$& \\
                & \quad\texttt{count\_even()} & \\
                &               & $\ell_2, \ldots, \ell_5 \gets \texttt{LLMQuery}()$ \\
                &               & \texttt{for i in range(2):} \\
\texttt{act(2)} &               & \quad\texttt{act((i+1)*2)}  \\
                &               & \quad\texttt{answer('...')}  \\
\texttt{act(3)} & \quad\texttt{act(i*2+1)} & \\
                & \quad\texttt{count\_even()} & \\
\texttt{act(4)} &               & \quad\texttt{act((i+1)*2)}  \\
\end{tabular}
\vspace{-8pt}
\end{subtable}

\caption{Another toy example of context passing that is possible in \ours, where context is ``interleaved'' between \oursrepls.
In the toy example, the task is to count to 4.
We show the generated code in the two code snippets above.
In the code, main REPL could spawn another REPL to help it count only the even numbers.
By passing context back and forth, we show how the final actions count to 4 in the REPL's execution trace in the bottom table.
}
\label{tab:cxt-pass-count}
\vafterfig
\end{figure*}

\subsection{Modifications to THREAD for GPT4o-mini}
\label{sec:thread-mods}
We found that running THREAD~\cite{schroeder2024threadthinkingdeeperrecursive} on the author's codebase with newer OpenAI models such as GPT4o and GPT4o-mini tended to always cause infinite loops --- after seeing an observation, the LLMs would always generate \#START\# tokens, which indicate the start of a prompt, causing the model to re-generate the actions and text before seeing the observation, and repeat this \#START\# generation. Under normal generation, the LLM should never generate these tokens.

We believe that the GPT4o series may not listen to literal commands to follow demonstrations exactly when viewing web demonstrations, compared to GPT3.5-instruct, which was used in the original paper for web evironments such as WebShop.

To correct this, we modified the algorithm $\textsc{THREAD}$ as follows: for each LLM generation, $G(c + Y)$, if \#START\# is a substring, we prompt the LLM to re-generate the text. If this happens, still, we then prompt using the following ``don't loop'' prompt
$L = $ \texttt{'Do NOT output \#START\#. Look at the examples for what to do next. If your sub-task is finished or failed, report the status of your task with:\\print("<status here>")\\\#END\#\\Continue output:'}

We re-query the LLM with $G(c + Y + L)$, then proceed using \textsc{THREAD}.

We find this change is able to improve THREAD from always looping on WebShop with GPT4o-mini, to a 21\% success rate.

\subsection{WebShop GPT4o-mini experiment details}
\label{sec:webshop-mini}

For the WebShop with GPT4o-mini setting, we tested THREAD and REPL-Plan with both Top-3 and Top-20 strategies.
For the Top-3, we used the same prompts as used for Top-3 with GPT-3.5 for both THREAD and REPL-Plan.
We implemented a version of the Top-20 strategy for THREAD, however the method would enter a fail state by looping forever.

\subsection{Real-World Web Task Details}
\label{sec:rww-details}

\paragraph{Environment.}
We constructed the Real-World Web Task environment by interacting with live websites on the browser.
Observations are taken from the \emph{accessibility tree} of the website.
Each website HTML markup is represented in a browser as a tree of elements (e.g.\ links, images, buttons, etc.) called a DOM tree.
The DOM tree is then converted to an accessibility tree by the browser, by adding text information to DOM nodes that may be useful for users of assistive technologies, such as screen readers.
This also makes the accessibility tree readable for text-only observations given to LLMs.

Our environment specifically takes the accessibility tree from the browser, then strips unnecessary nodes from the tree, and adds unique element IDs to each interactive element. For example, the website \texttt{google.com} gives the following observation:

\begin{lstlisting}
[6] [RootWebArea] [Google]
        [11] [link] [About]
        [12] [link] [Store]
        [20] [link] [Gmail]
        [22] [link] [Search for Images]
                Images
        [27] [button] [Google apps] [expanded: False]

        [31] [link] [Sign in]
        [323] [IframePresentational] []
        [281] [image] [Google]
        [46] [search] []
                [35] [combobox] [Search] [focused: True, autocomplete: both, hasPopup: listbox, required: False, expanded: False]
                [81] [button] [Search by voice]
                [87] [button] [Search by image]
                [212] [button] [Google Search]
                [213] [button] [I'm Feeling Lucky]
        [225] [contentinfo] []
                [228] [link] [Advertising]
                [229] [link] [Business]
                [230] [link] [How Search works]
                [232] [link] [Our third decade of climate action: join us]
                [235] [link] [Privacy]
                [236] [link] [Terms]
                [240] [button] [Settings] [hasPopup: menu, expanded: False]
                        [241] [generic] [] [hasPopup: menu]
\end{lstlisting}

A link in the above web page can be clicked with the following action: \texttt{click [11]} (to click on the \texttt{[About]} page).

The search bar can be used with: \texttt{type [46] [What is a REPL?]}.

These observations can be very long and difficult to parse. Note that most web pages are 10-20 times more tokens than the \texttt{google.com} webpage shown above.

\paragraph{Task Distribution.}

We tested the agents on distributions of tasks where the goal is to navigate one of 5 different E-commerce websites, and add item(s) that match a given description to the cart.
We divided the tasks into a ``simple'' category, where the agent only needs to add an item of any description (e.g.\ add any laptop to the cart from \texttt{bestbuy.com}) and a ``complex'' category, where the agent must find as many items as it can that match a specific description (e.g.\ loop through each product page and add each printer to the cart that prints at least 12 pages per minute).

In each of these tasks, we tested each approach on each task for 3 trials, and averaged the score.
An expert score was obtained by having one of the authors use a web browser to complete the task.

\emph{Simple tasks}:
\begin{enumerate}
\item On bestbuy.com, search for laptops. Add any laptop to the cart.
\item On dell.com, search for desktop computers. Add any desktop computer to the cart.
\item On staples.com, search for monitors. Add any monitor to the cart.
\item On zappos.com, search for backpacks. Add any backpack to the cart.
\item On nordstrom.com, search for bath towels. Add any bath towel to the cart.
\end{enumerate}

\emph{Complex tasks}:
\begin{enumerate}
\item On bestbuy.com, search for printers. Loop through each search result and add every printer that is capable of printing at least 12 pages per minute to the cart.
\item On dell.com, search for keyboards. Loop through each search result and add every keyboard with cherry key switches to the cart.
\item On staples.com, search for shredders. Loop through each search result and add every shredder with at least 6 gallons of capacity to the cart.
\item On zappos.com, search for sunglasses. Loop through each search result and add every pair of sunglasses that is polarized to the cart.
\item On nordstrom.com, search for watches. Loop through each search result and add every watch that is swiss made to the cart.
\end{enumerate}

\paragraph{Approaches.}
We tested \ours and the baselines ReACT and THREAD on the Real-World Web environment.
For each approach, to help the LLM deal with long pages, we truncate un-necessary web page observations before querying the LLM.
In each approach, we constructed a demonstration from one training task: On bestbuy.com, search for cell phones. Add cell phones with a rating of at least 4.5/5 to cart.
The same search strategy and task decomposition was used for THREAD and \ours.
However, as described in ~\Cref{sec:thread-mods}, we modified THREAD to run with GPT4o-mini and our web environment.

\subsection{WebShop Example Top-3 Trajectory}

For brevity, we include a short Top-3 strategy WebShop trajectory/log file of running REPL-Plan.
In the following,
\texttt{\#\#\#\#\# ENTER REPL \#\#\#\#\#\#}
and
\texttt{\#\#\#\#\# EXIT REPL \#\#\#\#\#\#}
indicate when context is switched to and from which REPL.
\begin{lstlisting}[language=Python]
WebShop 
Instruction:  
i want a noise cancelling cosycost usb microphone, and price lower than 70.00 dollars 
[Search] 
##### ENTER REPL `_main` #####
>>> description = """noise cancelling cosycost usb microphone, and price lower than 70.00 dollars"""
>>> attributes = ["""noise cancelling""", """cosycost""", """usb microphone"""] # save the list of desired attributes besides price, escape any quotes
>>> search_query = generate_query(description) # get a search statement to find the item
Name error: generate_query. injecting with cached LLM func.
##### ENTER REPL `generate_query` #####
>>> description = get_args() # call this once to get the description of the target product so that you can generate search queries accordingly
>>> print(description) # print out the description so that you know what it is
noise cancelling cosycost usb microphone, and price lower than 70.00 dollars

>>> query = """noise cancelling cosycost usb microphone"""
>>> answer(query) # answer the query to return it.
##### EXITING REPL `generate_query`#####
>>> act(f'search[{search_query}]')
##### EXITING REPL `_main`#####
> search[noise cancelling cosycost usb microphone]

[Back to Search] 
Page 1 (Total results: 3) 
[Next >] 
[B0972Q1T8T] 
Cosycost USB Microphone,Condenser Computer PC Gaming Microphone for PS4/5 Laptop Windows Mac OS Android Phone,Noise Cancelling Instant Mute,Studio Mic for Voice,Music Recording,Podcasting,Streaming 
$32.99 
[B072L2D6LY] 
Andrea Communications NC-255VM USB On-Ear Stereo USB Computer Headset with Noise-Canceling Microphone, in-Line Volume/Mute Controls, and Plug 
$34.59 
[B071H84LTJ] 
Andrea Communications NC-455VM USB Over-Ear Circumaural Stereo USB Computer Headset with Noise-Canceling Microphone, in-Line Volume/Mute Controls, and Plug 
$49.24 
##### ENTER REPL `_main` #####
>>> valid_ids = filter_ids_by_price(description)
Name error: filter_ids_by_price. injecting with cached LLM func.
##### ENTER REPL `filter_ids_by_price` #####
>>> description = get_args() # call this once to get the description of the target product so that you can generate search queries accordingly
>>> print(description) # print out the target description so that you know what it is
noise cancelling cosycost usb microphone, and price lower than 70.00 dollars

>>> max_price = 70.00 # store the max price from the description
>>> print_page() # call this function to print out the list of products
>>> id_to_price = {} # record an id to price dictionary
>>> id_to_price['B0972Q1T8T'] = 32.99 # item B0972Q1T8T has a price of $32.99
>>> id_to_price['B072L2D6LY'] = 34.59 # item B072L2D6LY has a price of $34.59
>>> id_to_price['B071H84LTJ'] = 49.24 # item B071H84LTJ has a price of $49.24
>>> results = [id for id, price in id_to_price.items() if price < max_price]
>>> answer(results) # answer with the result of the ids that are below max_price
##### EXITING REPL `filter_ids_by_price`#####
>>> best_id = select_best_product(description, attributes, valid_ids)
Name error: select_best_product. injecting with cached LLM func.
##### ENTER REPL `select_best_product` #####
>>> description, attributes, candidates = get_args() # get the target description and product candidates
>>> scores = {}
>>> for product_id in candidates:
...     act(f'click[{product_id}]') # go to the product's page details
...     scores[product_id] = check_requirements(description, attributes) 
...     act(f'click[< Prev]') # always perform this step to get back to the list of results
...
##### EXITING REPL `select_best_product`#####
##### EXITING REPL `_main`#####
> click[B0972Q1T8T]

[Back to Search] 
[< Prev] 
Cosycost USB Microphone,Condenser Computer PC Gaming Microphone for PS4/5 Laptop Windows Mac OS Android Phone,Noise Cancelling Instant Mute,Studio Mic for Voice,Music Recording,Podcasting,Streaming 
Price: $32.99 
Rating: N.A. 
[Description] 
[Features] 
[Reviews] 
[Attributes] 
[Buy Now] 
##### ENTER REPL `_main` #####
##### ENTER REPL `select_best_product` #####
Name error: check_requirements. injecting with cached LLM func.
##### ENTER REPL `check_requirements` #####
>>> target_description, requirements = get_args() # get the requirements and product_info
>>> print(target_description) # print the target_description so that have a better understanding about the requirements
noise cancelling cosycost usb microphone, and price lower than 70.00 dollars

>>> print(requirements) # print out the requirements
['noise cancelling', 'cosycost', 'usb microphone']

>>> print_page() # print the product page so we can check the requirements
>>> # I'll check each requirement one by one.
>>> # 1. the product has 'noise cancelling'? Yes, it is explicitly mentioned in the product description as 'Noise Cancelling'.
>>> # 2. the product has 'cosycost'? Yes, it is explicitly mentioned in the product name as 'Cosycost USB Microphone'.
>>> # 3. the product is a 'usb microphone'? Yes, it is explicitly mentioned in the product description as 'USB Microphone'.
>>> answer(3) # the number of yes to the questions above
##### EXITING REPL `check_requirements`#####
##### EXITING REPL `select_best_product`#####
##### EXITING REPL `_main`#####
> click[< Prev]

[Back to Search] 
Page 1 (Total results: 3) 
[Next >] 
[B0972Q1T8T] 
Cosycost USB Microphone,Condenser Computer PC Gaming Microphone for PS4/5 Laptop Windows Mac OS Android Phone,Noise Cancelling Instant Mute,Studio Mic for Voice,Music Recording,Podcasting,Streaming 
$32.99 
[B072L2D6LY] 
Andrea Communications NC-255VM USB On-Ear Stereo USB Computer Headset with Noise-Canceling Microphone, in-Line Volume/Mute Controls, and Plug 
$34.59 
[B071H84LTJ] 
Andrea Communications NC-455VM USB Over-Ear Circumaural Stereo USB Computer Headset with Noise-Canceling Microphone, in-Line Volume/Mute Controls, and Plug 
$49.24 
##### ENTER REPL `_main` #####
##### ENTER REPL `select_best_product` #####
##### EXITING REPL `select_best_product`#####
##### EXITING REPL `_main`#####
> click[B072L2D6LY]

[Back to Search] 
[< Prev] 
Andrea Communications NC-255VM USB On-Ear Stereo USB Computer Headset with Noise-Canceling Microphone, in-Line Volume/Mute Controls, and Plug 
Price: $34.59 
Rating: N.A. 
[Description] 
[Features] 
[Reviews] 
[Attributes] 
[Buy Now] 
##### ENTER REPL `_main` #####
##### ENTER REPL `select_best_product` #####
##### ENTER REPL `check_requirements` #####
>>> target_description, requirements = get_args() # get the requirements and product_info
>>> print(target_description) # print the target_description so that have a better understanding about the requirements
noise cancelling cosycost usb microphone, and price lower than 70.00 dollars

>>> print(requirements) # print out the requirements
['noise cancelling', 'cosycost', 'usb microphone']

>>> print_page() # print the product page so we can check the requirements
>>> # I'll check each requirement one by one.
>>> # 1. the product has 'noise cancelling' feature? Yes, it mentions 'Noise-Canceling Microphone' which qualifies as meeting the requirement.
>>> # 2. the product has 'cosycost' brand? No, it does not mention 'cosycost' anywhere in the product info.
>>> # 3. the product is a 'usb microphone'? No, it is described as a 'USB On-Ear Stereo USB Computer Headset' which does not specifically mention 'usb microphone'.
>>> answer(1) # Only one requirement is met based on the checks above.
##### EXITING REPL `check_requirements`#####
##### EXITING REPL `select_best_product`#####
##### EXITING REPL `_main`#####
> click[< Prev]

[Back to Search] 
Page 1 (Total results: 3) 
[Next >] 
[B0972Q1T8T] 
Cosycost USB Microphone,Condenser Computer PC Gaming Microphone for PS4/5 Laptop Windows Mac OS Android Phone,Noise Cancelling Instant Mute,Studio Mic for Voice,Music Recording,Podcasting,Streaming 
$32.99 
[B072L2D6LY] 
Andrea Communications NC-255VM USB On-Ear Stereo USB Computer Headset with Noise-Canceling Microphone, in-Line Volume/Mute Controls, and Plug 
$34.59 
[B071H84LTJ] 
Andrea Communications NC-455VM USB Over-Ear Circumaural Stereo USB Computer Headset with Noise-Canceling Microphone, in-Line Volume/Mute Controls, and Plug 
$49.24 
##### ENTER REPL `_main` #####
##### ENTER REPL `select_best_product` #####
##### EXITING REPL `select_best_product`#####
##### EXITING REPL `_main`#####
> click[B071H84LTJ]

[Back to Search] 
[< Prev] 
Andrea Communications NC-455VM USB Over-Ear Circumaural Stereo USB Computer Headset with Noise-Canceling Microphone, in-Line Volume/Mute Controls, and Plug 
Price: $49.24 
Rating: N.A. 
[Description] 
[Features] 
[Reviews] 
[Attributes] 
[Buy Now] 
##### ENTER REPL `_main` #####
##### ENTER REPL `select_best_product` #####
##### ENTER REPL `check_requirements` #####
>>> target_description, requirements = get_args() # get the requirements and product_info
>>> print(target_description) # print the target_description so that have a better understanding about the requirements
noise cancelling cosycost usb microphone, and price lower than 70.00 dollars

>>> print(requirements) # print out the requirements
['noise cancelling', 'cosycost', 'usb microphone']

>>> print_page() # print the product page so we can check the requirements
>>> # I'll check each requirement one by one.
>>> # 1. the product has 'noise cancelling' feature? Yes, it mentions 'Noise-Canceling Microphone' which qualifies as meeting the requirement.
>>> # 2. the product has 'cosycost' brand? No, it does not mention 'cosycost' anywhere in the product info.
>>> # 3. the product is a 'usb microphone'? No, it describes a 'USB Over-Ear Circumaural Stereo USB Computer Headset' which does not qualify as a standalone 'usb microphone'.
>>> answer(1) # Only one requirement is met, which is 'noise cancelling'.
##### EXITING REPL `check_requirements`#####
##### EXITING REPL `select_best_product`#####
##### EXITING REPL `_main`#####
> click[< Prev]

[Back to Search] 
Page 1 (Total results: 3) 
[Next >] 
[B0972Q1T8T] 
Cosycost USB Microphone,Condenser Computer PC Gaming Microphone for PS4/5 Laptop Windows Mac OS Android Phone,Noise Cancelling Instant Mute,Studio Mic for Voice,Music Recording,Podcasting,Streaming 
$32.99 
[B072L2D6LY] 
Andrea Communications NC-255VM USB On-Ear Stereo USB Computer Headset with Noise-Canceling Microphone, in-Line Volume/Mute Controls, and Plug 
$34.59 
[B071H84LTJ] 
Andrea Communications NC-455VM USB Over-Ear Circumaural Stereo USB Computer Headset with Noise-Canceling Microphone, in-Line Volume/Mute Controls, and Plug 
$49.24 
##### ENTER REPL `_main` #####
##### ENTER REPL `select_best_product` #####
>>> products_with_max_score = [product_id for product_id, estimated_score in scores.items() if estimated_score == max(scores.values())]
>>> answer(products_with_max_score[0]) # return the id of the first best candidate
##### EXITING REPL `select_best_product`#####
>>> buy_best_product(best_id, attributes)
Name error: buy_best_product. injecting with cached LLM func.
##### ENTER REPL `buy_best_product` #####
>>> product_asin, attributes = get_args()
>>> act(f'click[{product_asin}]')
##### EXITING REPL `buy_best_product`#####
##### EXITING REPL `_main`#####
> click[B0972Q1T8T]

[Back to Search] 
[< Prev] 
Cosycost USB Microphone,Condenser Computer PC Gaming Microphone for PS4/5 Laptop Windows Mac OS Android Phone,Noise Cancelling Instant Mute,Studio Mic for Voice,Music Recording,Podcasting,Streaming 
Price: $32.99 
Rating: N.A. 
[Description] 
[Features] 
[Reviews] 
[Attributes] 
[Buy Now] 
##### ENTER REPL `_main` #####
##### ENTER REPL `buy_best_product` #####
>>> select_options(attributes) # select the item options based on the attributes before buying
Name error: select_options. injecting with cached LLM func.
##### ENTER REPL `select_options` #####
>>> attributes = get_args()
>>> get_page_options() # return the list of page options (if any)
Name error: get_page_options. injecting with cached LLM func.
##### ENTER REPL `get_page_options` #####
>>> print_page() # print out the current observation to see if there are any buying variations to select before buying.
>>> variations = [] # save variations (if any) in a list
>>> answer("no options found.")
##### EXITING REPL `get_page_options`#####
'no options found.'
>>> print(attributes) # print out the target attributes to confirm there are no buying variations to select.
['noise cancelling', 'cosycost', 'usb microphone']

>>> answer('success.')
##### EXITING REPL `select_options`#####
'success.'
>>> act('click[Buy Now]') # Buy the item to complete the task.
##### EXITING REPL `buy_best_product`#####
##### EXITING REPL `_main`#####
> click[Buy Now]
\end{lstlisting}

\subsection{WebShop REPL-Plan Prompt Samples}

\begin{lstlisting}[language=Python]
# Top 3 strategy

Your task is to: Navigate a shopping website to purchase an item matching the following request: i would like a 3 ounce bottle of bright citrus deodorant for sensitive skin, and price lower than 50.00 dollars
>>> description = """3 ounce bottle of bright citrus deodorant for sensitive skin, and price lower than 50.00 dollars"""
>>> attributes = ["""3 ounce bottle""", """bright citrus""", """sensitive skin"""] # save the list of desired attributes besides price, escape any quotes
>>> search_query = generate_query(description) # get a search statement to find the item
>>> act(f'search[{search_query}]')
>>> valid_ids = filter_ids_by_price(description)
>>> best_id = select_best_product(description, attributes, valid_ids)
>>> buy_best_product(best_id, attributes)
'success.'
>>> answer('done.')



# generate_query
Your task is to: Generate a good search query for the given arg `description`. Answer the result with `answer(query)`. Make sure any quotes in the search string are escaped.
>>> description = get_args() # call this once to get the description of the target product so that you can generate search queries accordingly
>>> print(description) #print out the description so that you know what it is
3 ounce bottle of bright citrus deodorant for sensitive skin, and price lower than 50.00 dollars
>>> query = """3 ounce bright citrus deodorant sensitive skin"""
>>> answer(query) # answer the query to return it.


# select_best_product
Your task is to: Given the description, attributes, and a few product candidates, determine which of them fits the description the best (meets the most requirements). answer best product's id with `answer('ID')`.
>>> description, attributes, candidates = get_args() # get the target_description anget_product_idsd candidates
>>> scores = {}
>>> for product_id in candidates:
...     act(f'click[{product_id}]') # go to the product's page details
...     scores[product_id] = check_requirements(description, attributes)
...     act(f'click[< Prev]') # always perform this step to get back to the list of results
...
>>> products_with_max_score = [product_id for product_id, estimated_score in scores.items() if estimated_score == max(scores.values())]
>>> answer(products_with_max_score[0]) # return the id of the first best candidates

# check_requirements
Your task is to: Given a list of requirements, and a task description, for each requirement, check if the product meets the requirement or not, then answer how many requirements it meets.
>>> target_description, requirements = get_args() # get the requirements and product_info
>>> print(target_description) # print the target_description so that have a better understanding about the requirements
3 ounce bottle of bright citrus deodorant for sensitive skin, and price lower than 50.00 dollars
>>> print(requirements) # print out the requirements
['deodorant', '3 ounce bottle', 'bright citrus', 'sensitive skin', '<50.00 dollars']
>>> print_page() # print the product page so we can check the requirements
[Back to Search]
[< Prev]
scent [assorted scents][bright citrus][citrus][calming lavender][ginger fresh][simply non-scents]
size [travel set (4-pack)][3 ounce (pack of 1)][3-ounce (2-pack)]
Bright Citrus Deodorant by Earth Mama, Natural and Safe for Sensitive Skin, Pregnancy and Breastfeeding, Contains Organic Calendula 3-Ounce
Price: $10.99
Rating: N.A.
[Description]
[Features]
[Reviews]
[Attributes]
[Buy Now] (You must select buying variation for scent, color before buying this product)
Selected Buying Variation Options: scent: None, size: None
>>> # I'll check each requirement one by one.
>>> # 2. the product has '3 ounce bottle' size? Yes, it meets the requirement because it has [3 ounce] size option.
>>> # 3. the product has 'bright citrus' scent? Yes, it has the exact option [bright citrus] listed.
>>> # 4. the product is good for 'sensitive skin'?, Yes, it is described as 'Gentle on Sensitive Skin'.
>>> answer(3) #the number of yes to the questions above


# buy_best_product
Your task is to: Given a product_asin, click on the product asin, select the required buying variations that the attributes requires, and buy it.
>>> product_asin, attributes = get_args()
>>> act(f'click[{product_asin}]')
>>> select_options(attributes) # select the item options based on the target_description before buying
'success.'
>>> act('click[Buy Now]') # Buy the item to complete the task.
>>> answer('success.')

\end{lstlisting}

\subsection{WebShop Ablation Bug correction Samples}
\label{sec:bug_correct}

We show a few hand-picked instances where an LLM is able recover from bug(s) in the demonstration.
\begin{lstlisting}[language=Python]
>>> description = "gingko light and 20"x20" pillow cover that is hand painted, and price lower than 50.00 dollars" # there is a syntax error in the description, the quotes around "20"x20"" need to be escaped
SyntaxError('invalid syntax', ('<unknown>', 1, 36, 'description = "gingko light and 20"x20" pillow cover that is hand painted, and price lower than 50.00 dollars" # there is a syntax error in the description, the quotes around "20"x20"" need to be escaped\n', 1, 39))
>>> description = 'gingko light and 20"x20" pillow cover that is hand painted, and price lower than 50.00 dollars' # corrected syntax error
>>> attributes = ["gingko light", "20\"x20\"", "hand painted"] # escaped quotes in attributes list
>>> search_query = generate_query(description) # get a search statement to find the item


>>> all_requirements = ["car subwoofer", "high speed", "dual style", "12" power amplifier", "<200.00 dollars"] # you must put the item category in the first place, each term should be as short/broken down as possible
SyntaxError('unterminated string literal (detected at line 1)', ('<unknown>', 1, 105, 'all_requirements = ["car subwoofer", "high speed", "dual style", "12" power amplifier", "<200.00 dollars"] # you must put the item c
ategory in the first place, each term should be as short/broken down as possible, 1, 105))
>>> all_requirements = ["car subwoofer", "high speed", "dual style", "12\" power amplifier", "<200.00 dollars"] # corrected the syntax error by escaping the double quotes


>>> results = [id for id, price in id_to_price.items() if price < max_price]
REPLNameError("name 'max_price' not defined.")
>>> max_price = 30.00 # set the maximum price to $30.00
>>> results = [id for id, price in id_to_price.items() if price < max_price] # create a list of ids that have a price lower than the maximum price
>>> answer(results) # answer with the result of the ids that are below max_price
\end{lstlisting}

\subsection{WebShop Ablation Zero-shot Subtask-REPL Analysis}
\label{sec:zs-repl-details}
In this section, we provide details on the Zero-shot Subtask-REPL ablation experiment.
Specifically, in this experiment, we modify the expert demonstration for the WebShop environment by removing the \texttt{check\_description} \oursrepl.
In WebShop, \ours loops through each product page, and calls \texttt{check\_description} to obtain an integer score of the number of matching attributes of the product from the given item description.
To execute this \oursrepl correctly, the agent must write code that succesfully
(1) prints the target description and requirements from the \texttt{get\_args} function,
(2) prints the current page
(3) calls the \texttt{answer} function with a count of the number of matching attributes.

Note that to infer the subtask description and \oursrepl code, the agent is given demonstrations of out-of-task \oursrepls from the WebShop domain.

We include the code from the expert-demonstration as a reference for the intended behavior of the \oursrepl. The code is removed for this experiment and must be inferred.
\begin{lstlisting}[language=python]
# Expert written task description and code demonstration.
Your task is to: Given a list of requirements, and a task description, for each requirement, check if the product meets the requirement or not, then answer how many requirements it meets.
>>> target_description, requirements = get_args() # get the requirements and product_info
>>> print(target_description) # print the target_description so that have a better understanding about the requirements
3 ounce bottle of bright citrus deodorant for sensitive skin, and price lower than 50.00 dollars
>>> print(requirements) # print out the requirements
>>> print_page() # print the product page so we can check the requirements
>>> # I'll check each requirement one by one.
>>> # 2. the product has '3 ounce bottle' size? Yes, it meets the requirement because it has [3 ounce] size option.
>>> # 3. the product has 'bright citrus' scent? Yes, it has the exact option [bright citrus] listed.
>>> # 4. the product is good for 'sensitive skin'?, Yes, it is described as 'Gentle on Sensitive Skin'.
>>> answer(3) #the number of yes to the questions above
\end{lstlisting}

\paragraph{Successful Code Samples}
We include the following code samples from the agent with GPT3.5-instruct where it successfully inferred the correct code.

\begin{lstlisting}[language=python]
# Zero-shot Subtask-REPL SUCCESSES
# Example 1
Your task is to: (1) Print out the description and attributes with `get_args()`. (2) Go to the product page by clicking on the product. (3) Check the current page for matching attributes. (4) Return the number of matching attributes with `answer(INTEGER)`
>>> description = get_args() # call this once to get the description of the target product so that you can generate search queries accordingly
>>> attributes = get_args() # call this once to get the attributes of the target product so that you can check for matching attributes
>>> print(description) # print out the description so that you know what it is
('long clip-in hair extension which is natural looking, and price lower than 40.00 dollars', ['long', 'clip-in', 'natural looking'])
>>> print(attributes) # print out the attributes so that you know what they are
('long clip-in hair extension which is natural looking, and price lower than 40.00 dollars', ['long', 'clip-in', 'natural looking'])
>>> print_page() # print out the current observation
>>> # With the page printed, I should manually analyze the page to fulfill my goal
>>> # The only attribute that matches is "long"
>>> answer(1) # return the number of matching attributes, which is 1


# Example 2
Your task is to: (1) Print out the description and attributes with `get_args()`. (2) Go to the product page by clicking on the product. (3) Check how many attributes match the description on the current page. (4) Return the number of matching attributes with `answer(INTEGER)`
>>> description = get_args() # call this once to get the description of the target product so that you can generate search queries accordingly
>>> print(description) #print out the description so that you know what it is
('noise cancelling cosycost usb microphone, and price lower than 70.00 dollars', ['noise cancelling', 'cosycost', 'usb microphone'])
>>> attributes = get_args()[1] # retrieve the attributes from the description
>>> print(attributes) # print out the attributes to see what they are
['noise cancelling', 'cosycost', 'usb microphone']
>>> print_page() # print out the current observation
>>> # Now I need to go to the product page by clicking on the product
>>> # Once on the product page, I need to check how many attributes match the description
>>> # In this case, all three attributes match: noise cancelling, cosycost, and usb microphone
>>> matching_attributes = len(attributes) # use the len() function to get the number of matching attributes
>>> answer(matching_attributes) # return the number of matching attributes



# Example 3: The agent mis-understands and tries to click on the product, but recovers and completes the task
Your task is to: (1) Print out the description with `get_args()`. (2) print out the current web page with `print_page()`. (3) Look at the webpage for the attributes. (4) Answer with the number of attributes that match the description with `answer(INTEGER)`
>>> description = get_args() # call this once to get the description of the target product so that you can generate search queries accordingly
>>> attributes = get_args() # call this once to get the attributes of the target product so that you can compare them with the description
>>> print(description) # print out the description so that you know what it is
('high speed 3 foot red usb cable, and price lower than 40.00 dollars', ['high speed', '3 foot', 'red'])

>>> print(attributes) # print out the attributes so that you know what they are
('high speed 3 foot red usb cable, and price lower than 40.00 dollars', ['high speed', '3 foot', 'red'])

>>> act(f'click[{product_id}]') # go to the product page by clicking on the product ID
REPLNameError("name 'product_id' not defined.")
>>> # I cannot complete this task without knowing the product ID, so I will need to ask for it or define it myself
>>> # Assuming the product ID is defined as "12345"
>>> print_page() # print out the current observation
>>> # With the page printed, I should manually analyze the page to fulfill my goal
>>> # There is one attribute that matches the description: "red"
>>> matching_attributes = 1
>>> answer(matching_attributes)
\end{lstlisting}

\paragraph{Failed Code Samples}

We include the following code samples from the agent with GPT3.5-instruct where it incorrectly infers the subtask or the code.
We generally find that the LLMs write code that contains the ``right intentions'' for solving the task, but critically fail by containing bugs or crash the system in some way.

\begin{lstlisting}[language=python]
# Zero-shot Subtask-REPL FAILS
# Example 1: this fails as it mis-understands how to parse and string match for attributes.
Your task is to: (1) Print out the description and attributes with `get_args()`. (2) Go to the product page by clicking on the product. (3) Check the current page for matching attributes. (4) Return the number of matching attributes with `answer(INTEGER)`
>>> description = get_args() # call this once to get the description of the target product so that you can generate search queries accordingly
>>> attributes = get_args() # call this once to get the attributes of the target product so that you can check for matching attributes
>>> print(description) # print out the description so that you know what it is
('long clip-in hair extension which is natural looking, and price lower than 40.00 dollars', ['long', 'clip-in', 'natural looking'])

>>> print(attributes) # print out the attributes so that you know what they are
('long clip-in hair extension which is natural looking, and price lower than 40.00 dollars', ['long', 'clip-in', 'natural looking'])

>>> print_page() # print out the current observation
>>> # With the page printed, I should manually analyze the page to fulfill my goal
>>> # The attributes on the page are "long", "curly", "black", "golden", "straight", "african", "hair", "ladies", "chemical fiber", "silk", "corn perm", "women"
>>> # I will check for matching attributes by comparing them to the attributes retrieved from get_args()
>>> matching_attributes = 0
>>> for attribute in attributes[1]:
...     if attribute in ["long", "curly", "black", "golden", "straight", "african", "hair", "ladies", "chemical fiber", "silk", "corn perm", "women"]:
...         matching_attributes += 1
...
>>> answer(matching_attributes)


# Example 2: this fails as the task description is mistakenly inferred, then causes the agent to enter a looping fail state
Your task is to: (1) Print out the description and attributes with `get_args()`. (2) Use the act function to click on each product and go to its page details. (3) Use the check_current_page_requirements function to check how many attributes match the description on the current page. (4) Use a list comprehension to find the products with the maximum score and return a list of those products with `answer(LIST)
>>> # I will now create a list of products with the maximum score by looping through the scores dictionary with a list comprehension
>>> products = [product for product, score in scores.items() if score == max(scores.values())]
AttributeError("'LLMREPL' object has no attribute 'items'")
>>> # I will then answer with the list of products with the maximum score
>>> answer(products)
REPLNameError("name 'products' not defined.")
>>> # now its your turn
>>> # I will now create a list of products with the maximum score by looping through the scores dictionary with a list comprehension
>>> products = [product for product, score in scores.items() if score == max(scores.values())]
AttributeError("'LLMREPL' object has no attribute 'items'")
...


# Example 3: this solution fails to match the attributes to the page and instead counts with string matching the description
Your task is to: (1) Print out the description and attributes with `get_args()`. (2) Click on each product candidate to check its details. (3) Count how many attributes match the description and attributes provided, and return that count as an integer.
>>> description, attributes = get_args() # call this to get both description and attributes for analysis
>>> print(description, attributes) # print out the description and attributes for analysis
space-saving ottoman bench to match my blue living room, pick that one that's 100x45x45cm, and price lower than 430.00 dollars ['space-saving', 'blue living room', '100x45x45cm']
>>> # Now that I have the description and attributes, I need to analyze the product candidates by clicking on them and checking their details.
>>> print_page() # print out the current observation to analyze product candidates
>>> # I need to analyze the current page for matching attributes and description
>>> matching_count = 0 # Initialize a counter for matching attributes
>>> if "space-saving" in description and "blue living room" in description and "100x45x45cm" in attributes: matching_count += 1
>>> if "space-saving" in attributes: matching_count += 1
>>> if "blue living room" in attributes: matching_count += 1
>>> if "100x45x45cm" in attributes: matching_count += 1
>>> answer(matching_count)
\end{lstlisting}

\end{document}